\documentclass[11pt]{article}

\usepackage[final]{acl}

\usepackage{times}
\usepackage{latexsym}

\usepackage[T1]{fontenc}

\usepackage[utf8]{inputenc}

\usepackage{microtype}

\usepackage{inconsolata}
\usepackage{multirow}
\usepackage{tabularx}
\usepackage{graphicx}
\usepackage{subcaption}

\usepackage{placeins}

\usepackage{booktabs}
\usepackage{multirow}
\usepackage{xcolor}
\usepackage{colortbl}

\usepackage{paralist}
\usepackage{booktabs}   
\usepackage{multirow}   
\usepackage{makecell}   
\definecolor{asqaYellow}{RGB}{255, 253, 220}
\definecolor{eli5Green}{RGB}{220, 245, 220}
\definecolor{expertAgua}{RGB}{180, 254, 220}

\usepackage{todonotes}
\makeatletter
\newcommand*\iftodonotes{\if@todonotes@disabled\expan
dafter\@secondoftwo\else\expandafter\@firstoftwo\fi} 
\makeatother

\definecolor{lilac}{RGB}{200, 162, 200}

\title{Explicit Evidence Grounding via Structured Inline Citation Generation}

\author{%
  Anar Yeginbergen\textsuperscript{1} \quad
  Amelie Wührl\textsuperscript{2} \quad
  Anna Rogers\textsuperscript{2} \quad
  Rodrigo Agerri \textsuperscript{1} \\[6pt]
  \textsuperscript{1}University of the Basque Country (UPV/EHU) \qquad
  \textsuperscript{2}IT University of Copenhagen \\[6pt]
}

\begin{document}
\maketitle
\begin{abstract}

As AI systems become more widely adopted, the demand for factual and faithful generation grows. Properly attributing information through citations becomes, therefore, crucial. This work introduces FullCite, a framework that, in contrast to most previous works, generates structured inline citations linking each claim to both its source document and supporting evidence. FullCite proposes three strategies to inline citation generation: prompt-based generation, constrained decoding over a citation grammar, and posthoc span alignment. Using three question answering benchmarks, namely, ASQA, BioASQ, and ExpertQA, we assess citation quality and faithfulness along three dimensions: document-level correctness, evidence span identification, and claim-citation faithfulness. Our evaluation shows that while LLMs are generally effective at identifying relevant documents, they struggle to identify the precise supporting spans within them. This gap suggests that achieving faithful attributed QA will require research to place greater emphasis on precise evidence span identification.

\end{abstract}

\section{Introduction}

Given the widespread adoption of AI systems such as ChatGPT in our daily lives, large language models (LLMs) are increasingly becoming a common alternative to traditional search engines, providing direct answers rather than ranked lists of related sources \cite{nakano2021webgpt, shi2025know}.

In particular, in high-stakes domains (medical, scientific, legal, etc.) both users and practitioners need to verify that generated statements are grounded in trustworthy evidence, rather than relying solely on the LLM's parametric knowledge \cite{schreieder2025attribution}.


\begin{figure}[t]
    \centering
    \includegraphics[width=0.9\linewidth]{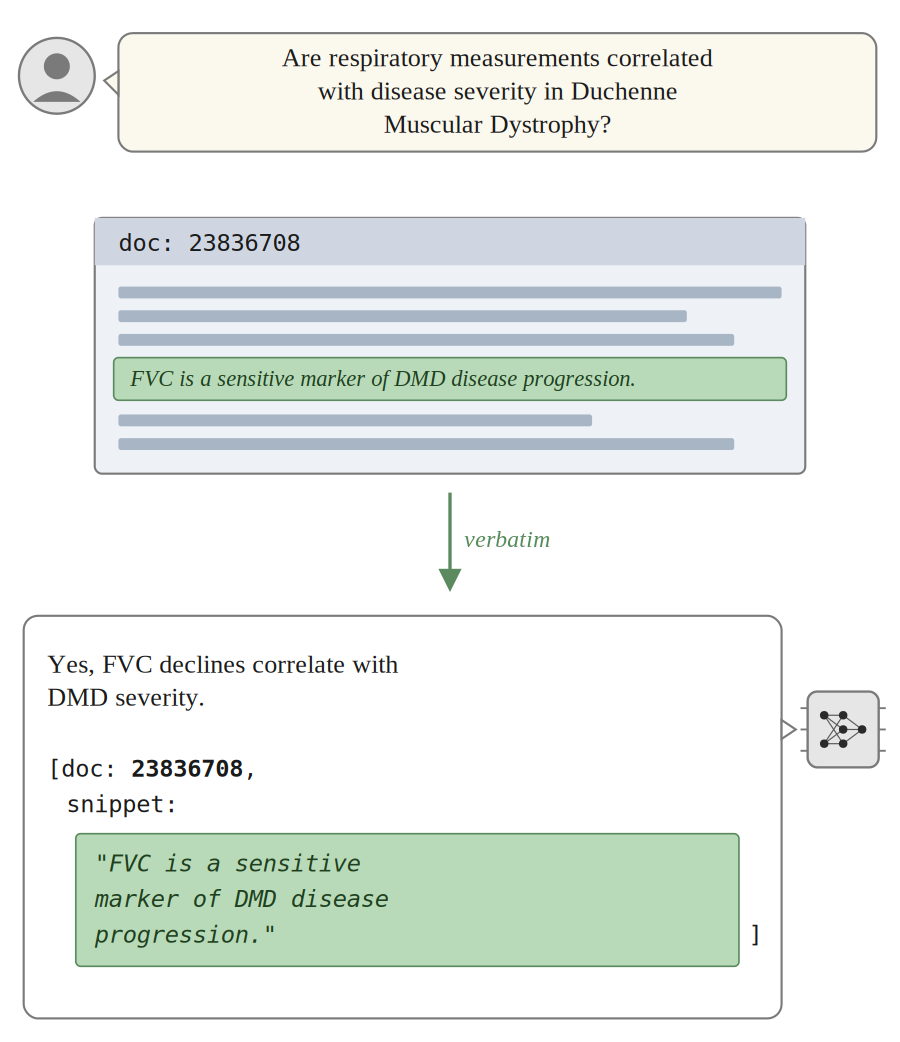}
    \caption{An overview of FullCite, structured inline generation.}
    \label{fig:intro}
\end{figure}


We believe that combining both document- and fine-grained evidence-level citations at the same time provides a more reliable and transparent text attribution. Motivated by this, we introduce \textit{FullCite}, a framework for generating both document-level and evidence-level citations simultaneously for long-context question-answering (QA). An overview of \textit{FullCite} is shown in Figure \ref{fig:intro}.

\begin{table*}[ht]

    \centering
    \begin{tabular}{c|cccc}
        Dataset & Domain & Question Types & N. of Questions & Documents (avg.) \\
        \hline
        BioASQ & Bio-medical & yes-no/fact./summ./list  & 177/175/156/170 & 5.0 \\
        ASQA & General & fact. & 580 & 1.61 \\
        ExpertQA & 32 domains & yes-no/fact./summ./list  & 259/702/314/244  & 3.25 \\
    \end{tabular}
    \caption{Data distribution across 3 datasets: BioASQ, ASQA and ExpertQA. \textit{N. of Questions} shows number of questions according to \textit{Question Types} in the dataset. \textit{Documents (avg.)} shows the number of context documents on average used for citation generation. An example of questions for each quesiton types are in Appendix \ref{appendix:data-qtype-example}.}
    \label{tab:data}
\end{table*}
In \textit{FullCite}, we analyze different strategies for data attribution in the long-context scenarios using three QA benchmarks: BioASQ \cite{tsatsaronis2015overview, krithara2023bioasq}, ExpertQA \cite{malaviya23expertqa} and ASQA \cite{stelmakh2022asqa}. 
In fact, we show that jointly referencing documents and supporting evidence spans leads to more transparent and faithful grounding, rather than relying on either document references or evidence snippets alone. 
These are the main contributions:
\noindent \textbf{(C1)}: Our results show that LLMs perform better in document-level citations, while still struggling to consistently identify the correct supporting evidence.

\noindent \textbf{(C2)}: For span-based evidence identification, FullCite provides three different citation strategies: prompt-based generation, constrained decoding via a finite-state automaton over the citation grammar, and posthoc evidence span alignment, and shows that posthoc yields the largest gains in correct evidence identification by increasing snippet-F1 from 12.80 to 61.87 for ASQA.

\noindent \textbf{(C3)}: We discover two systematic patterns undermining the evidence grounding process: (i) strong primacy bias in document selection, i.e., 81.8\% of BioASQ citations target only the first two of five context documents, consistent with \textit{lost-in-the-middle} phenomenon \cite{liu-etal-2024-lost}, and (ii) citation omission on binary yes/no questions, which inflates baseline scores in ways that vanish once attribution is enforced.



The key strength of FullCite lies in its joint coverage: it is the only framework that simultaneously optimizes for both document-level and evidence span attribution while maintaining competitive semantic similarity scores, making it the most balanced and transparent approach to faithful attributed QA.

\section{Related Work}

Attributed answer generation is an established research problem \cite{nakano2021webgpt, bohnet2022attributed}, made more prominent by the widespread use of LLMs for information access. 

Retrieval-Augmented Generation (RAG) has emerged as a method that grounds model outputs in externally retrieved documents, allowing the model to generate answers conditioned on relevant contextual information \cite{NEURIPS2020_6b493230, guu2020retrieval}. RAG improves LLMs performance \cite{gao2023retrieval} by incorporating external knowledge. 
However, the presence of retrieved documents does not guarantee that the model faithfully uses them during generation \cite{wallat2024correctness, zhou2023context}. Models may still generate answers based on memorized knowledge \cite{longpre2021entity, xu2024knowledge}, incorrectly attribute claims to irrelevant documents, or produce outputs that are only partially supported by the retrieved evidence.

\begin{figure*}[ht]
    \centering
    \includegraphics[width=1\linewidth]{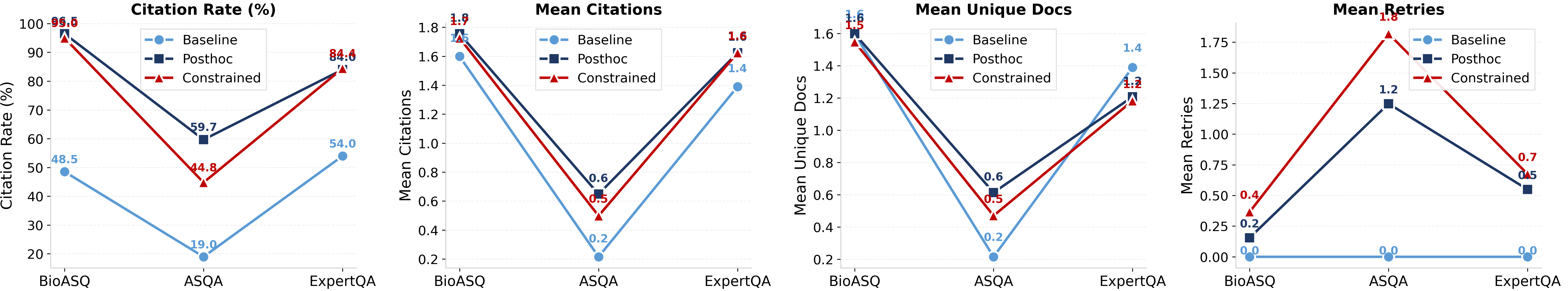}
    \caption{Overall quality comparison of citations across settings and datasets.}
    \label{fig:overall-quality}
\end{figure*}
To address this, prior work has explored explicit citation generation, where LLMs are prompted to reference the sources used to answer a question \cite{menick2022teaching, nakano2021webgpt, bohnet2022attributed, gao2023enabling}.

However, document-level citations alone provide coarse-grained attributions. A referenced document may contain related information, but lack any supporting evidence \cite{huang2024learning, cao2024verifiable}. 

To address this, fine-grained citation approaches have been proposed that reference specific evidence spans within source documents rather than entire documents \cite{huang-chang-2024-citation, cao2024verifiable, xu2025aliice}. Existing approaches generally follow one of three strategies: (i) generating citations directly from the model’s parametric knowledge through prompting \cite{sun2022recitation, huang-chang-2024-citation}, (ii) retrieving documents and conditioning answer generation on the retrieved context \cite{NEURIPS2020_6b493230, bohnet2022attributed, gao2023enabling}, or (iii) generating answers first and retrieving supporting evidence later \cite{gao2023rarr, schreieder2025attribution}.

\section{Data}

To assess inline citation generation across domains and question types, we perform our experiments on three well-known datasets well-suited for the inline attribution problem and contain annotated gold documents to answer the input question:

\noindent \textbf{BioASQ} \cite{krithara2023bioasq}: a biomedical question-answering dataset that, along with the various fine-grained annotations, includes the gold documents from PubMed and the corresponding evidence location of the correct answer. 

\noindent \textbf{ASQA} \cite{stelmakh2022asqa}: a dataset of factoid long-form ambiguous question answering enhanced with related knowledge passages from Wikipedia.

\noindent \textbf{ExpertQA} \cite{malaviya23expertqa}: a dataset designed specifically for factuality and attribution evaluation tasks that spans across 32 different domains, including medicine, with expert-annotated knowledge sources for each question.

We further apply the following adaptations to the datasets. 
Although ASQA and ExpertQA are designed for data attribution, they provide only document-level annotations, identifying the source document but not the exact evidence spans needed for fine-grained citation. We therefore apply an additional annotation step to extract atomic evidence spans from the gold documents using GPT-5.4-mini \cite{singh2025openai}.
To validate this procedure, we apply it to BioASQ, which provides both document- and evidence-level annotations, and compare the extracted spans against the gold annotations. The results confirm strong alignment, achieving over 90\% ROUGE-L and chrF++ \citep{lin2004rouge} scores and 85\% F1 token overlap across all metrics.

We manually review and refine the extracted evidence for each question in ExpertQA and ASQA, finding that not all retrieved documents are suitable for fine-grained citation extraction. Specifically, many documents are informational or introductory in nature, providing general background on the topic rather than explicit supporting statements (see Appendix \ref{appendix:skipped-question-example} for an example). While such documents may implicitly guide a model toward the correct answer, they do not lend themselves to evaluating a model's explicit citation capabilities. We therefore retain only documents where supporting evidence is directly and unambiguously present in the retrieved context, discarding approximately 500 examples from ExpertQA and 350 from ASQA.

Table~\ref{tab:data} summarizes the statistics of each dataset's final version, with examples provided in Appendix~\ref{appendix:data-qtype-example}. BioASQ and ExpertQA cover four question types: yes/no, factoid, list, and summary; ASQA includes factoid questions only.

\begin{table*}[ht]
\centering
\small
\setlength{\tabcolsep}{2pt}
\newcolumntype{Y}{>{\columncolor{asqaYellow}}c}
\newcolumntype{G}{>{\columncolor{eli5Green}}c}
\newcolumntype{P}{>{\columncolor{expertAgua}}c}
\begin{tabular}{l YYY GGG PPP}
\toprule
\multirow{3}{*}{Model}
  & \multicolumn{3}{c}{ASQA}
  & \multicolumn{3}{c}{BioASQ}
  & \multicolumn{3}{c}{ExpertQA} \\
\cmidrule(lr){2-4} \cmidrule(lr){5-7} \cmidrule(lr){8-10}
  & \cellcolor{white}Doc-F1 & \cellcolor{white}Snippet-F1 & \cellcolor{white}Similarity
  & \cellcolor{white}Doc-F1 & \cellcolor{white}Snippet-F1 & \cellcolor{white}Similarity
  & \cellcolor{white}Doc-F1 & \cellcolor{white}Snippet-F1 & \cellcolor{white}Similarity \\
\midrule
\multicolumn{10}{c}{\textit{Prompt-based}} \\
\midrule
Qwen3-8B           & 33.87 & 12.80 & 56.55 & 58.08 &  6.18 & 64.89 & 56.42 & 5.56 & 64.61 \\
Gemma-3-12B-it      & 18.16 & 12.42 & 63.56 & 36.57 & 28.84 & 69.43 & 42.15 & 16.01 & \textbf{69.16} \\
\midrule
\multicolumn{10}{c}{\textit{Posthoc}} \\
\midrule
Generate-then-Retrieve            & \textbf{93.74} & \textbf{75.07} & 42.93 & 47.36 & 16.83 & 57.29 & \textbf{82.50} & 32.70 & 42.41 \\
FullCite (Qwen3-8B)           & 80.98 & 61.87 & 52.17 & \textbf{49.25} & 24.23 & 56.75 & 53.92 & 28.44 & 56.82 \\
FullCite (Gemma-3-12B-it)    & 53.17 & 41.80 & 73.46 & 43.37 & 20.90 & 71.89 & 44.18 & 30.13 & 69.02 \\
\midrule
\multicolumn{10}{c}{\textit{Constrained}} \\
\midrule
ReClaim (Qwen3-8B)            & - & 58.22 & 68.07 & - & \textbf{43.96} & 68.01 & - & \textbf{33.55} & 61.71  \\
ReClaim (Gemma-3-12B-it)            & - & 42.16 & \textbf{73.85} & - & 10.31 & \textbf{78.35} & - & 13.12 & 68.53  \\
FullCite (Qwen3-8B)           & 74.59 & 55.11 & 51.60 & 43.37 & 17.35 & 53.53 & 65.80 & 27.23 & 53.54 \\
FullCite (Gemma-3-12B-it)     & 39.43 & 29.99 & 72.16 & 38.91 & 20.76 & 76.12 & 56.06 & 27.34 & 67.99 \\
\bottomrule
\end{tabular}
\caption{Results across ASQA, BioASQ, and ExpertQA for three citation strategies: prompt-based, posthoc, and constrained decoding. We report Doc-F1 (document-level citation accuracy), Snippet-F1 (evidence span identification), and Similarity (semantic faithfulness of cited spans to the generated answer). Bold indicates the best score per metric and dataset.}
\label{tab:main_results}
\end{table*}

\section{Experimental Setup}

We aim to understand if the structured verbatim inline citations approach, \textit{FullCite}, facilitates reliable attribution for RAG-based QA while comparing it with popular state-of-the-art methods. To measure attribution quality, we evaluate the citations at the document and snippet level and compare it with other methods. 

\subsection{\textit{FullCite}: Verbatim Inline Citation Generation}
\textit{FullCite} requires that every claim in the answer is followed by a citation comprising (i) the identifier of the source document, and (ii) a verbatim evidence span from the selected document that supports the preceding claim, in the form of \texttt{\{doc\_id: <document identifier>, snippet: <verbatim text>\}}. We test three variants that differ in how structure is enforced.

\textbf{Prompt-based Citation Generation.} The model is instructed to provide verbatim evidence citations from the context documents after every generated claim. The compliance with the format and verbatim grounding relies entirely on the model’s instruction-following ability without any decoding-time intervention or post-processing.

\textbf{Constrained Citation Generation. }
We enforce citation format at inference time with a logic processor via a finite-state automaton over the citation grammar. To guarantee the structure and verbatim evidence spans,  the automaton tracks whether the model is currently generating a claim, a document ID, or a snippet, and at each state restricts the next token to those consistent with the grammar, and with the verbatim content from the context document at evidence generation time. In case of failure at any state, the model tries again up to three times from the beginning with an increased temperature by 0.5 after each failure.

\textbf{Posthoc Citation Generation with Approximation.} 
Generating an attributed text using constrained decoding is efficient compared to prompt-based citation generation and guarantees that the output is verbatim and follows the exactly defined structure. However, this setting could be too strict, and if the model fails to follow the structure, the provided output is discarded as a valid output. After analyzing results from the experiments from baseline and constrained decoding settings, we noticed that LLMs often produce near-verbatim snippets that differ from the source document by a small number of tokens and require further processing. 

Henceforth, we introduce a third citation strategy where we generate text at inference time, and given the generated text, we try to approximately find the part of the document that is cited. In other words, in case if the generated citation snippet does not match word by word one or several words, the system tries to reconstruct it by finding the most similar snippet from the document text, given that the document ID is generated correctly. We select the evidence from the document by computing word-level overlap Jaccard similarity \cite{niwattanakul2013using}. We empirically find that setting the similarity score to 0.7 (see Appendix \ref{appendix: similarity}) gives best results and use it for all the settings and models. An example of when posthoc is advantageous over constrained is illustrated in Appendix \ref{appendix:snippet-example}.

\textbf{Implementation details.} We perform experiments using two open-weight LLMs of comparable scale from different model families: Qwen3-8B \cite{qwen3technicalreport} and Gemma3-12b-it \cite{team2024gemma}. All settings were performed and evaluated under the same default hyperparameter settings of starting temperature of 0.7, top-p of 0.95, top-k of 50, with output length not exceeding 1500 tokens. Snippet length is limited to be between 20 and 512 characters, to avoid trivially short chunks, while accommodating the length of the whole document. 

\subsection{Baselines}

We compare \textit{FullCite} against two prior methods that represent two strategies for citation generation: posthoc retrieval-based attribution and inline constrained-decoding attribution.

\textbf{Generate-then-retrieve} \cite{gao2023enabling, zhang2025longcite, wang2025medcite} a posthoc attribution strategy in which the main idea is to first generate an answer to the given question with LLMs and then retrieve supporting documents from the knowledge base. This strategy is used for coarse-level attribution, i.e., citing only the document. We further adapt this method to evidence-level citation. We first generate the answer to the question using Gemma-12b-it. We split the outputs and the relevant pre-retrieved documents into sentences. Next, we use BM25 \cite{robertson2009probabilistic} and \textit{all-MiniLM-L6-v2}\footnote{\url{https://huggingface.co/sentence-transformers/all-MiniLM-L6-v2}} for retrieving relevant evidence from the document.

\textbf{ReClaim} \cite{xia2025ground}. The inline citation generation, the main idea of this method is to provide citations after every claim that is generated through constrained decoding. This method works in two passes, and each one requires training the separate model for the task: first, generate the claim and second, evidence from the context document. However, ReClaim is only focused on evidence-level attribution without the strict necessity of document attribution. Including document citation returned diminishing results. Therefore we only report evidence-level citation results.


\subsection{Evaluation}
We evaluate generated citations along four axes: 

\textbf{Document-level evaluation}    
measures if the cited document is correct.
From each prediction, we extract a set of unique cited document IDs and compute document-level F1 score against the set of gold annotated documents in the dataset. These metrics show the coverage of sources independent of the evidence snippet. 

\textbf{Snippet-level evaluation}
measures the correctness of the cited evidence span within the documents.
One of the central questions is how well models cite \emph{verbatim} snippets from the source. Therefore, we evaluate how closely each generated evidence snippet matches gold-annotated snippet at the surface level, using three string-overlap metrics: ROUGE-L \cite{lin2004rouge} for longest common subsequence overlap, Jaccard F1 \cite{niwattanakul2013using} for token-set overlap and chrF++ \cite{popovic2017chrf++} for character-level n-gram overlap. 

\textbf{Claim-citation faithfulness} evaluates the semantic relation between the generated claim and citation using similarity metrics, LLM-as-a-judge and human evaluation. While document and span-level evaluation measure whether the generated citation is in fact the \textit{right} one, this assesses whether the citation faithfully \textit{supports} the claim it follows. We measure faithfulness in three ways: 

\begin{compactitem}
  \item We use sentence transformers \cite{reimers-gurevych-2019-sentence} to compute a similarity between a claim and a citation. We use \textit{all-MiniLM-L6-v2}.
  \item We employ OpenAI's GPT-5.4\footnote{\url{https://developers.openai.com/api/docs/models/gpt-5.4}} model as LLM-as-a-Judge to measure the support and relatedness levels between claim and citation. We instruct the model to answer two questions and output a scale from 1-5 as an answer for the evaluation: (Q1) whether the citation supports the claim, and (Q2) whether the citation and the document are related to the claim. We interpret the results as \textit{No Support} if a score of 1 or 2 is given, \textit{Partial Support} for the score of 3, and \textit{Full Support} for the scores 4 and 5.
  \item We conduct a manual evaluation identical to the evaluation for LLM-as-a-Judge. Two authors independently annotate a sample of 50 examples to evaluate the support level and relatedness between the claim and citation.
\end{compactitem}

\textbf{Downstream question answering.}
 We run the experiments on the downstream question-answering task citation-free to compare the influence of the generated output on downstream answer correctness. We report the results based on the downstream task for each question type in the three datasets: the macro F1 score based on \textit{yes/no questions} and ROUGE-L score for the rest of the question types.


\begin{figure}
    \centering
    \includegraphics[width=\linewidth]{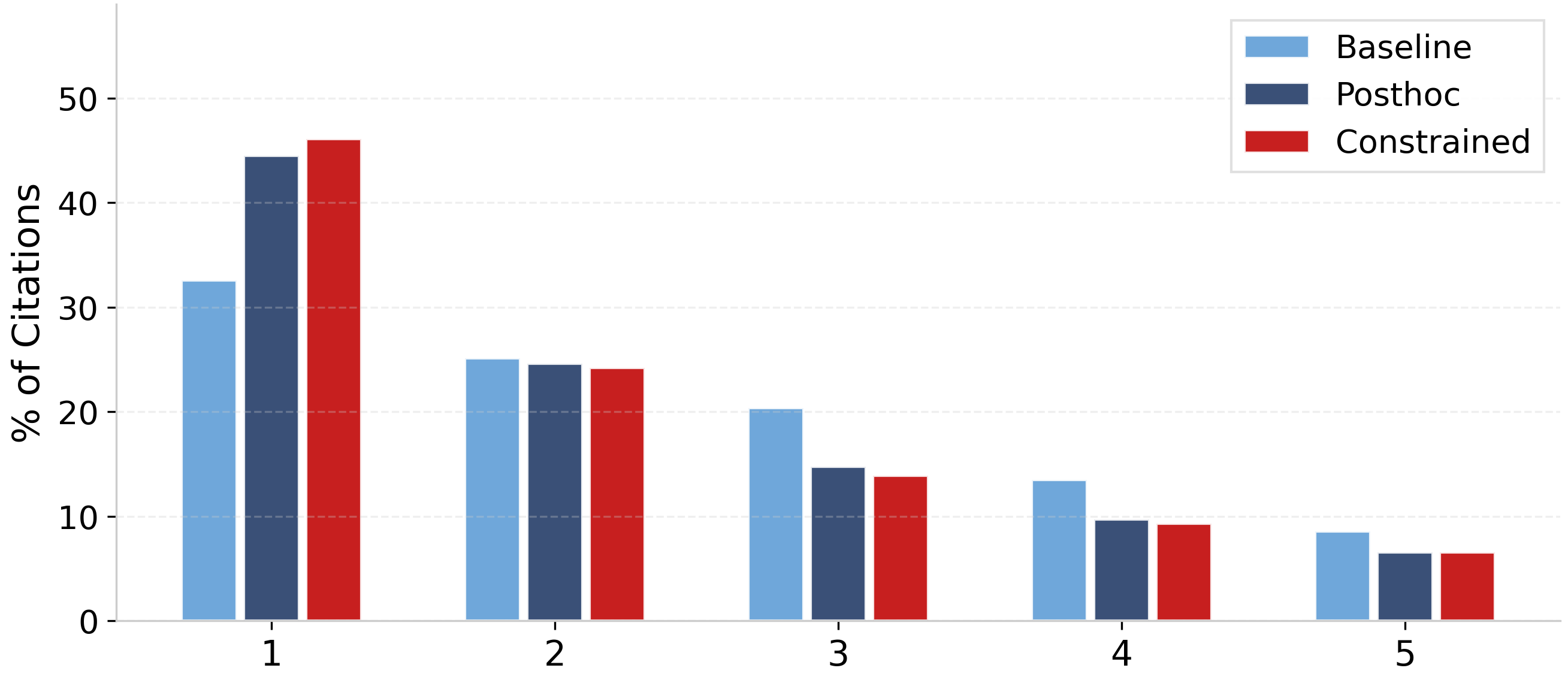}
    \caption{Position of the cited document in BioASQ.}
    \label{fig:doc_position}
\end{figure}

\section{Results}

\textbf{Can LLMs cite the right documents and establish the exact textual evidence that supports their answers?} We analyze the reliability of FullCite, the faithfulness of citations with respect to the claims of model predictions using both similarity and human evaluations.

\vspace{0.3cm}
\noindent \textit{1. How reliably does FullCite reference relevant documents and snippets from the evidence documents?} 
Table \ref{tab:main_results} shows the results. The models we evaluate reliably identify relevant documents, as reflected in consistently high Doc-F1 scores.  However, they struggle to localize the precise evidence span within those documents, which is evident from the substantially lower Snippet-F1 scores. The posthoc \textit{Generate-then-retrieve} (baseline) method exhibits the highest Doc-F1 score, reaching 80 and 94\% in ASQA and ExpertQA, whereas snippet-level localization is lower, with 43.96 with ReClaim being the highest for BioASQ and 33.55 for ExpertQA. 

In the \textit{posthoc} setting of FullCite, where generated citations are aligned to the closest matching passages, yields substantial gains over the baseline for ASQA with the gains from 12.80 to 61.87 with Qwen3-8B and from 12.42 to 41.80 with Gemma-12B. For BioASQ and ExpertQA, the effects differ by model: with Qwen, document-level F1 declines from 58.08 to 49.25 while span-level performance improves from 6.18 to 24.23, trading document coverage for span precision. Gemma exhibits the opposite pattern: document F1 increases to 43.37 compared to 36.57 in the baseline, and snippet-F1 drops from 28.84 in the baseline to 20.90 in the posthoc setting.
Under the \textit{constrained} generation setting, performance gains are more modest compared to those observed in the \textit{posthoc} approach.

\vspace{0.3cm}
\noindent \textit{2. How faithful are citations to the claims in model responses?}
\begin{figure}[ht]
    \centering
    \includegraphics[width=0.9\linewidth]{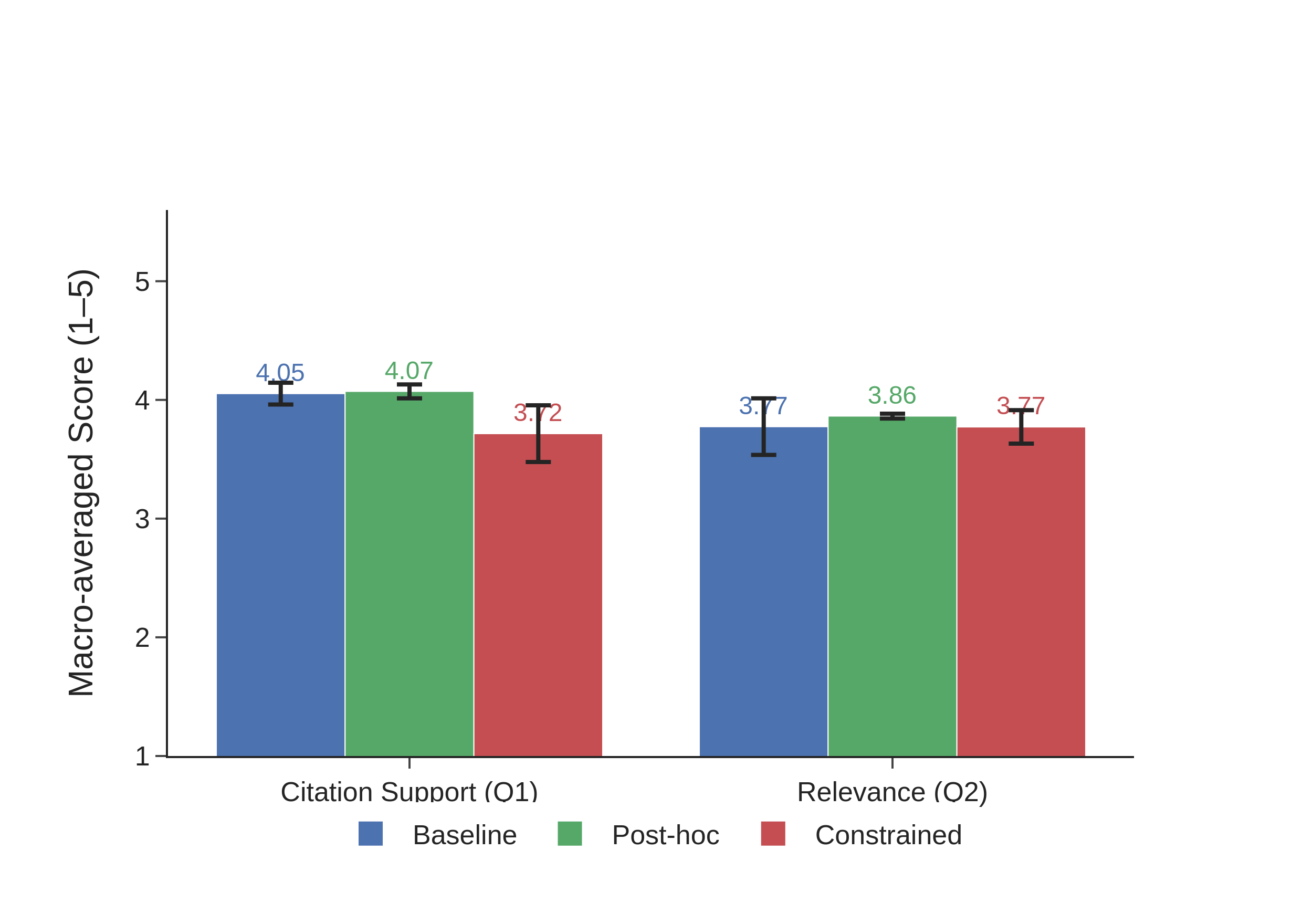}
    \caption{Average scores for Citation Support (Q1) and Citation Relevance (Q2) by GPT-5.4.}
    \label{fig:llm-aggregated-avg}
\end{figure}
We use semantic similarity as a proxy for faithfulness, measuring the degree to which each cited span is semantically aligned with the claim it supports. As shown in Table~\ref{tab:main_results}, the two models exhibit quite a different behaviour. \textit{Qwen3-8B} achieves its highest similarity scores under the FullCite baseline setting across all datasets, with values declining under both the posthoc and constrained configurations. \textit{Gemma-3-12B-it}, by contrast, proves more stable: similarity scores are largely preserved or even improved when moving to the posthoc and constrained settings. On ASQA, the posthoc setting yields the largest gain, rising by approximately 10 points from 63.56 (prompt-based) to 73.46; on BioASQ under the constrained setting, the score improves by nearly 7 points, from 69.43 to 76.12. For ExpertQA, similarity remains relatively stable across all three settings, fluctuating within a narrow 1.2-point range.


Overall, both the \textit{posthoc} and \textit{constrained} strategies improve evidence identification, yet this comes at the cost of reduced semantic alignment between the selected citations and their associated claims. In the case of \textit{Generate-then-Retrieve}, citations are selected posthoc by ranking candidate spans according to similarity score; however, the highest-scoring span does not necessarily constitute the most faithful evidence for the generated claim, as reflected in its comparatively lower similarity scores relative to the other methods.



\begin{figure}[ht]
    \centering
    \includegraphics[width=0.82\linewidth]{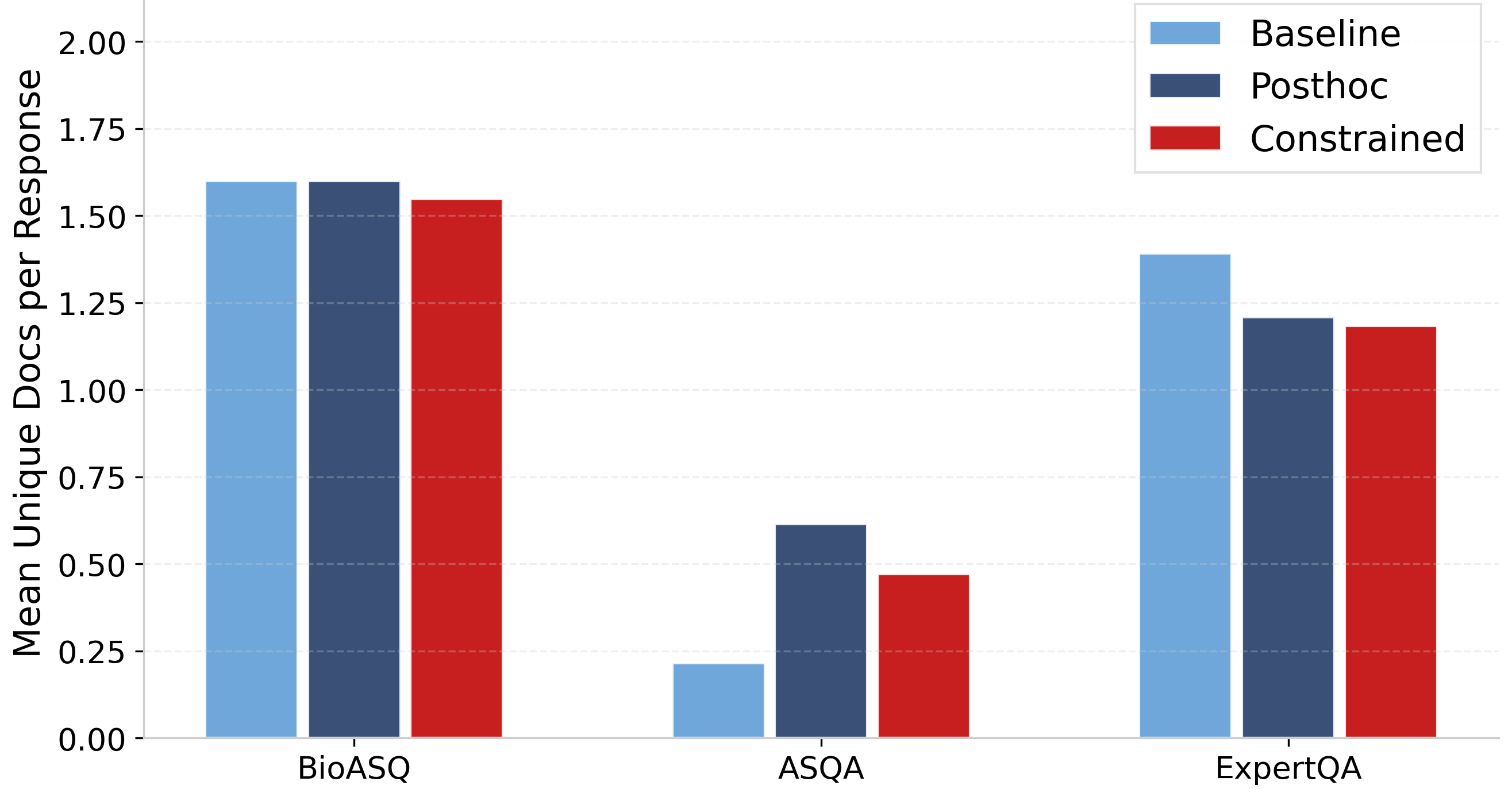}
    \caption{Ratio of unique cited documents.}
    \label{fig:unique-docs}
\end{figure}

\begin{figure}
    \centering
    \includegraphics[width=\linewidth]{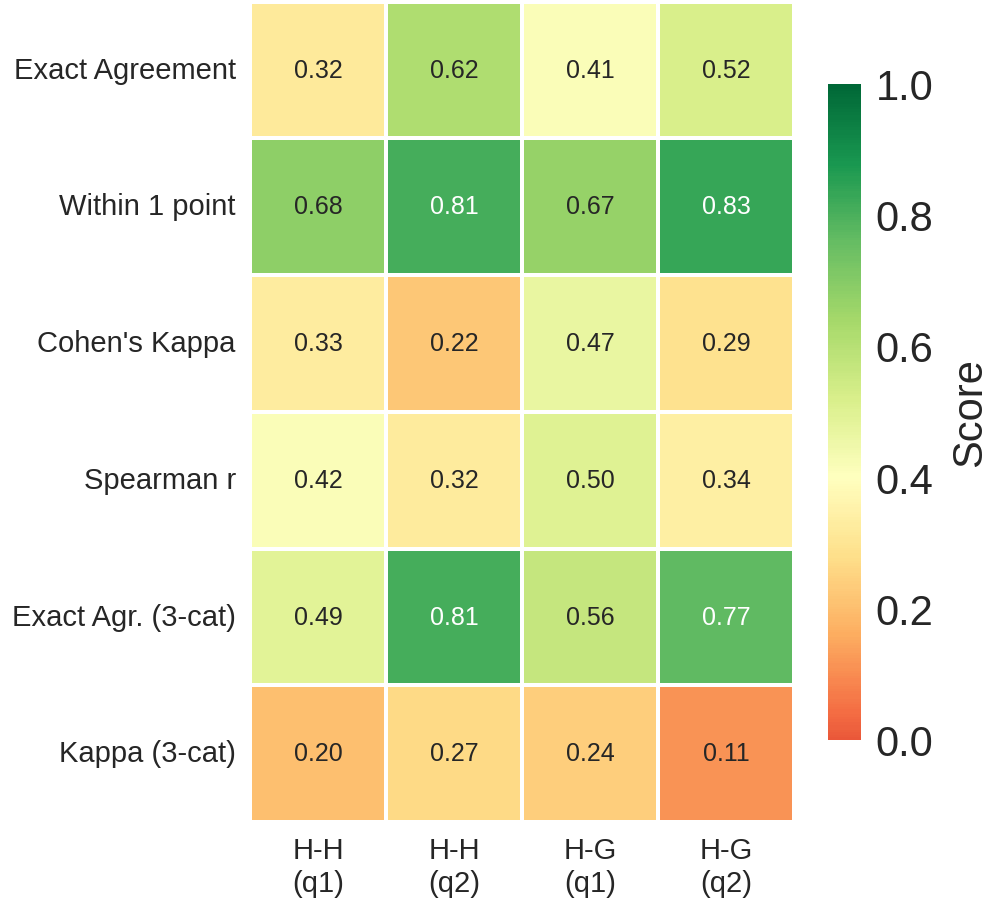}
    \caption{Agreement and correlation between human annotators and GPT-5.4. "H" corresponds to \textit{Human} and "G" corresponds to \textit{GPT-5.4}, "q1" corresponds to the Support question and "q2" to the Relevance question.}
    \label{fig:human-llm-agreemnt}
\end{figure}

\noindent \textit{3. LLM and Human evaluation.} We use LLM-as-a-Judge and human evaluation to evaluate the same set of examples in order to assess how LLMs align or diverge in the citation evaluation task by asking two questions: (Q1): \textit{Does the citation support the claim that is being cited?} and (Q2) \textit{Is the citation relevant to the claim?}. The answer is provided on a scale between 1 and 5. The interpretation of each score is in Appendix \ref{appendix:rubrics}. Along with individual scores, we interpret the results as \textit{No Support/Relevance} scores 1 and 2,  \textit{Partial Support/Relevance} score 3 and \textit{Full Support/Relevance} scores 4 and 5. We evaluate the citations generated by \textit{Gemma-12b-it} using the posthoc \textit{FullCite} strategy.  We report the agreement and correlation between human and LLM in Figure \ref{fig:human-llm-agreemnt}.

The most frequent source of disagreement between human-human and human-LLM pairs involves adjacent scores, namely, annotators assigning 1 where the other assigns 2, and vice versa. We therefore report results under both strict exact agreement and a relaxed within-1-point criterion. Following the consolidation into three categories (\textit{No}, \textit{Partial}, and \textit{Full} support/relevance), we run the same evaluation across all three datasets and citation strategies. As shown in Figure~\ref{fig:llm-aggregated-avg}, all methods consistently obtain high \textit{Support} scores and slightly lower \textit{Relevance} scores on average, with the \textit{posthoc} strategy achieving the strongest performance. The full score distributions by dataset and method are provided in Appendix~\ref{appendix:llm-eval-scatter}.

\textbf{What is the most effective strategy for grounding LLM-generated answers in precise span-based evidence?}

\noindent \textit{1. Quality of correct verbatim evidence.} When it comes to evaluating how verbatim the correct are we evaluate the character and token-level overlap and ROUGE-L. From Table \ref{tab:verbatim}, Appendix \ref{appendix:verbatim}, we can see that most of the time \textit{FullCite} exhibits the highest correct and verbatim span citations across all settings, outperforming \textit{ReClaim} by at least 20\% in Overlap and ROUGE-L, and by around 45 points in chrF++.


\vspace{0.3cm}
\noindent \textit{2. How does FullCite compare to other methods?} To situate \textit{FullCite} against existing approaches, we include two reference methods in our main results (Table \ref{tab:main_results}): \textit{Generate-then-Retrieve}, a representative \textit{Posthoc} baseline, and \textit{ReClaim}, a representative \textit{Constrained} method. The main takeaway is that FullCite offers the best joint coverage. In other words, it is the only framework that reports both Doc-F1 and Snippet-F1 while maintaining competitive similarity scores, making it the most balanced approach for faithful attributed QA.

\begin{figure}
    \centering
    \includegraphics[width=0.8\linewidth]{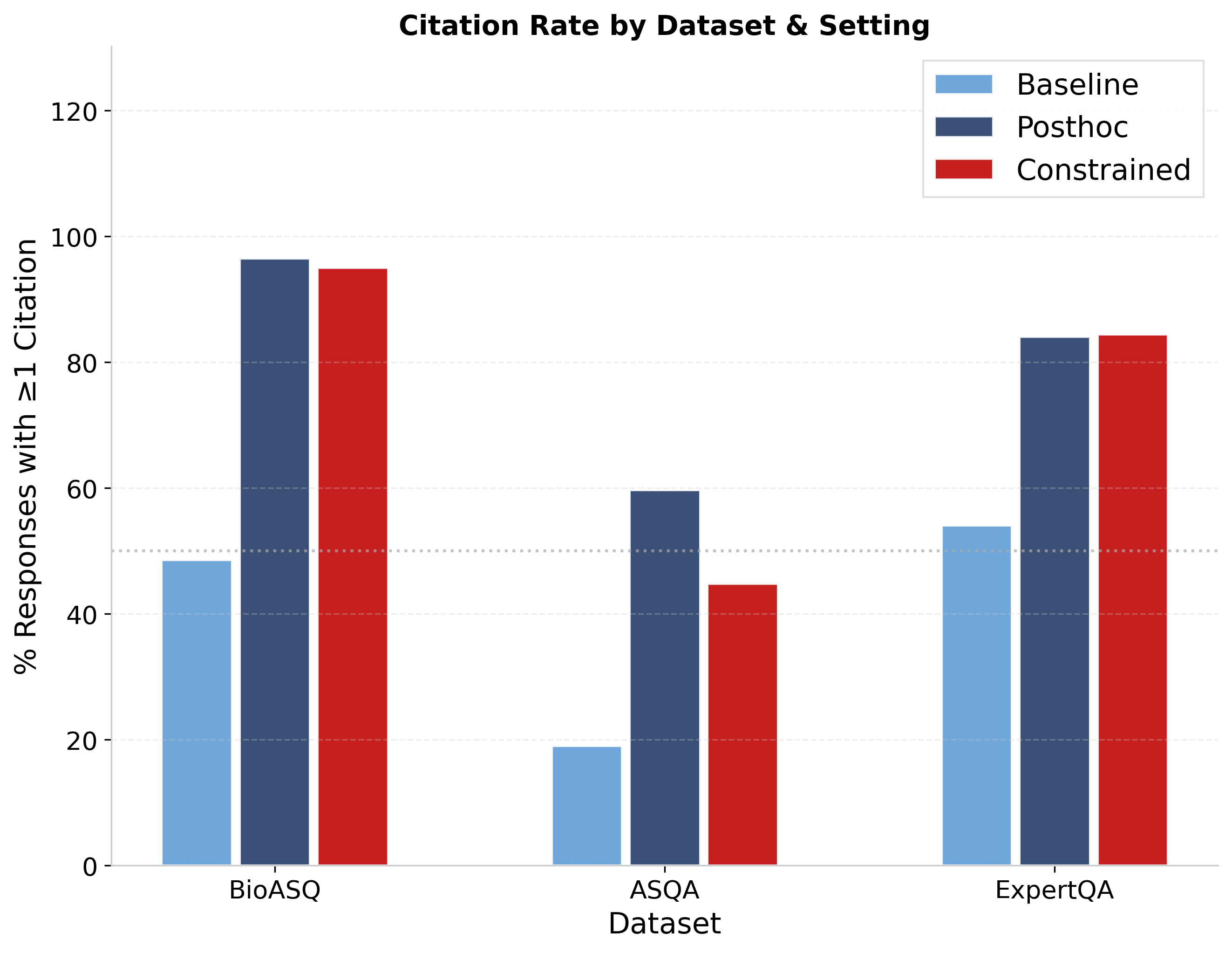}
    \caption{Overall citation rate per method.}
    \label{fig:citation_rate}
\end{figure}

\begin{table*}[ht]
\centering
\small
\setlength{\tabcolsep}{3pt}
\newcolumntype{Y}{>{\columncolor{asqaYellow}}c}
\newcolumntype{G}{>{\columncolor{eli5Green}}c}
\newcolumntype{P}{>{\columncolor{expertAgua}}c}
\begin{tabular}{l c cccc cccc}
\toprule
\multirow{1}{*}{Model}
  & \multicolumn{1}{c}{ASQA}
  & \multicolumn{4}{c}{BioASQ}
  & \multicolumn{4}{c}{ExpertQA} \\
\cmidrule(lr){2-2} \cmidrule(lr){3-6} \cmidrule(lr){7-10}
  & \cellcolor{white}Factoid
  & \cellcolor{white}Yes/No & \cellcolor{white}Factoid & \cellcolor{white}List & \cellcolor{white}Summary
  & \cellcolor{white}Yes/No & \cellcolor{white}Factoid & \cellcolor{white}List & \cellcolor{white}Summary \\
\midrule
w/out citation  & 12.49 & 38.61 & 36.74 & 30.67 & 27.01 & 49.02& 19.32 & 17.31 & 17.51          \\
\midrule
Baseline       & 26.66 & 81.77 & 42.22 & 41.79 & 38.67 & 60.68 & 36.25 & 17.71 & 17.55  \\
Posthoc        & 15.34 & 69.47 & 52.63 & 48.05 & 42.29 & 48.39 & 15.72  & 20.18 & 17.28 \\
Constrained    & 12.50 & 63.88  & 52.30 & 46.24 & 39.94 & 38.52 & 15.45  & 19.67 & 16.58 \\
\bottomrule
\end{tabular}
\caption{Performance comparison on downstream task with citation and without: ASQA, BioASQ, and ExpertQA. For \textit{Yes/No} questions we report macro-F1 and for the rest, we report ROUGE-L.}
\label{tab:downstream}
\end{table*}

\section{Analysis}
\label{sec: error}
We conduct a detailed analysis of the models' citation `behavior' in each \textit{FullCite} setting (\textit{Baseline}, \textit{Constrained}, \textit{Posthoc}). Figure \ref{fig:overall-quality} illustrates how, on average, each setting performs in citation rate, unique docs, and mean retries (for Posthoc and Constrained). From this observation, we can conclude that \textit{Posthoc} leads in all these dimensions, followed by \textit{Constrained} decoding strategy.

\noindent \textbf{Number of citations per question.}
Figure \ref{fig:citation_rate} illustrates the average number of generated citations per question for each setting and dataset. Combined with the results illustrated in Table \ref{tab:main_results}, it is evident that both structure-enforced settings of \textit{FullCite} produce not only more frequent but also better attribution 
compared to the \textit{baseline} setting. 

In the Baseline settings, around 13\% of the time, LLM correctly identifies the document but fails to provide a supporting snippet in BioASQ, 5.3 and 6.6\% in the case of ASQA and ExpertQA, respectively, which does not occur in other settings (more details in Appendix \ref{appendix: error_statistics}).

\noindent \textbf{Citations across question types.} 
Yes/No questions have the lowest citation rates across all datasets and settings (see Appendix \ref{appendix: citation_rate_qtype}). The models tend to answer with a direct affirmative or negative response without providing any grounding in a document, especially in \textit{Baseline}. However, citations occur later on, after one or two ungrounded sentences and since we limit the generation to 1500 tokens, the models have fewer tokens for more citations.  

\noindent \textbf{Citation coverage.} 
As shown in the low unique document rate in Figure \ref{fig:overall-quality} and Figure \ref{fig:citation_rate}, models consistently fail to cite all relevant gold documents.
Even with constrained decoding, the majority of responses cite only a few documents, instead of the full set. This is most pronounced in \textit{Baseline} for ASQA but improves over \textit{posthoc} and \textit{constrained} settings. 
Moreover, primacy bias in selecting document for citation is evident. In case of BioASQ, which always holds five documents as context, in 81.8\% of the time cite the first two documents only, as shown in Figure \ref{fig:doc_position}. This is consistent with the \textit{lost-in-the-middle} \cite{liu-etal-2024-lost} phenomenon.

\textbf{Structurally invalid citations.} 
In case of invalid structure and attribution in \textit{posthoc} and \textit{constrained} citation, the model runs up to three retries. If all fail, the model returns \textit{``Cannot answer using provided documents."}. ASQA suffers the most from failed retries (around 40\% of the time), suggesting that the documents may not contain explicit evidence for models to be able to generate verbatim (see Appendix \ref{appendix: error_statistics}). An example of such questions is illustrated in Appendix \ref{appendix:skipped-question-example}. Although we exclude the questions with unattributable context documents,
some of the context documents still remain challenging for LLMs.
Additionally, invalid citations occur when the generated evidence spans are verbose and do not align with the gold documents. This happens in 35 examples of ASQA in \textit{Constrained} setting. More details in Appendix \ref{appendix: error_statistics}, Table \ref{tab:citation-behavior}.

\paragraph{Performance on downstream task.}

Table~\ref{tab:downstream} reports downstream task performance across question types. ExpertQA and BioASQ cover four types, \textit{yes/no}, \textit{factoid}, \textit{summary}, and \textit{list}, evaluated with macro F1 and ROUGE-L, respectively. The \textit{posthoc} FullCite setting yields the strongest overall results, with BioASQ showing the largest average gain of 19.85 points. Although the prompt-based baseline scores highest on \textit{yes/no} questions, Section~\ref{sec: error} shows this is an artifact of models omitting citations altogether for this question type. Since our task is attributed QA rather than plain QA, a correct but ungrounded answer is not enough; the baseline's higher \textit{yes/no} scores thus reflect a citation coverage failure rather than good performance, and \textit{posthoc} and \textit{constrained} settings are precisely designed to address this.

For \textit{factoid} and \textit{summary} questions, performance increases from 36.7 to 52.6\% on BioASQ under the posthoc setting, with lower gains on ASQA and ExpertQA. Improvements are most consistent for \textit{list} questions across both posthoc and constrained settings. Overall, citation grounding proves most beneficial for short, verifiable answers rather than free-form long-form responses. Citation rates per question type and setting are provided in Appendix~\ref{appendix: citation_rate_qtype}.

\section{Conclusion}

This work studies attributed question answering under three citation
paradigms \textit{Prompt-based}, \textit{Posthoc}
(citations approximated and fixed after prompt-based generation), and \textit{Constrained}
(generation restricted to structure and context vocabulary) and introduced
\textit{FullCite}, a generation of structured citations after every claim at inference-time. Experiments on ASQA, BioASQ, and ExpertQA with
two open-weight LLMs (Qwen3-8B and Gemma-3-12B-it) assessed citations
along three axes: document selection (Doc-F1), span localization
(Snippet-F1), and claim-citation faithfulness. 

Results show that document-level identification appears to be the easier sub-task with Doc-F1 substantially exceeding Snippet-F1 across all settings and datasets, indicating that span localization remains the principal source of difficulty. We also observe a trade-off between snippet-level localization and claim-citation faithfulness. We interpret this trade-off with caution: across all settings, claim and citation are generated jointly, which is likely to inflate their semantic similarity independently of whether the citation would substantiate the claim
against the source. Therefore, the \textit{Generate-then-retrieve} method yields high document-level performance but fails in evidence localization, especially in domain-specific questions. 

A similar caveat applies to the Baseline's apparent advantage on yes/no questions, which, on closer inspection, reflects omitted citations rather than better grounding. We show through human and LLM evaluation that inline citation generation yields high support and relevance scores between claim and citation. Taken together, the findings indicate that faithful attributed QA may depend less on document-level retrieval than on evidence-level grounding and on evaluation that can distinguish surface similarity from genuine support.

\section*{Limitations}


FullCite's evaluation is conditioned by several methodological choices. The three benchmarks used, ASQA, BioASQ, and ExpertQA, cover general, biomedical, and multi-domain questions, but findings may not transfer to other high-stakes domains such as legal or financial QA. Similarly, experiments are conducted with only two open-weight models (Qwen3-8B and Gemma-3-12B-it), leaving open the question of whether the observed patterns hold for larger models, proprietary systems, or models explicitly trained for RAG-style attribution. The automatic metrics employed, Doc-F1, Snippet-F1, and semantic similarity, provide useful proxies but do not fully capture whether a cited span genuinely entails the associated claim; our human evaluation partially addresses this gap, yet remains limited in scale.

Beyond evaluation scope, two systematic phenomena constrain FullCite's current reach. The strong primacy bias, 81.8\% of BioASQ citations targeting only the first two of five context documents, suggests that LLMs fail to reliably process full retrieved contexts, a limitation FullCite does not explicitly mitigate. Likewise, the tendency of LLMs to omit citations entirely on yes/no questions remains an open challenge: while the posthoc and constrained settings are designed to enforce attribution, neither offers a principled mechanism for grounding binary answers. More generally, fine-tuning or instruction-tuning models explicitly for joint document- and span-level citation generation was not explored, and may provide trade-offs beyond what prompt-based or decoding-time strategies can achieve.

Nevertheless, the central claim of this work stands: FullCite is the only framework that jointly addresses document-level and evidence span attribution without sacrificing semantic faithfulness.


\bibliography{custom}

\appendix



\section{Citation quality under different similarity thresholds}
\label{appendix: similarity}
\begin{figure}[h]
    \centering
    \includegraphics[width=1\linewidth]{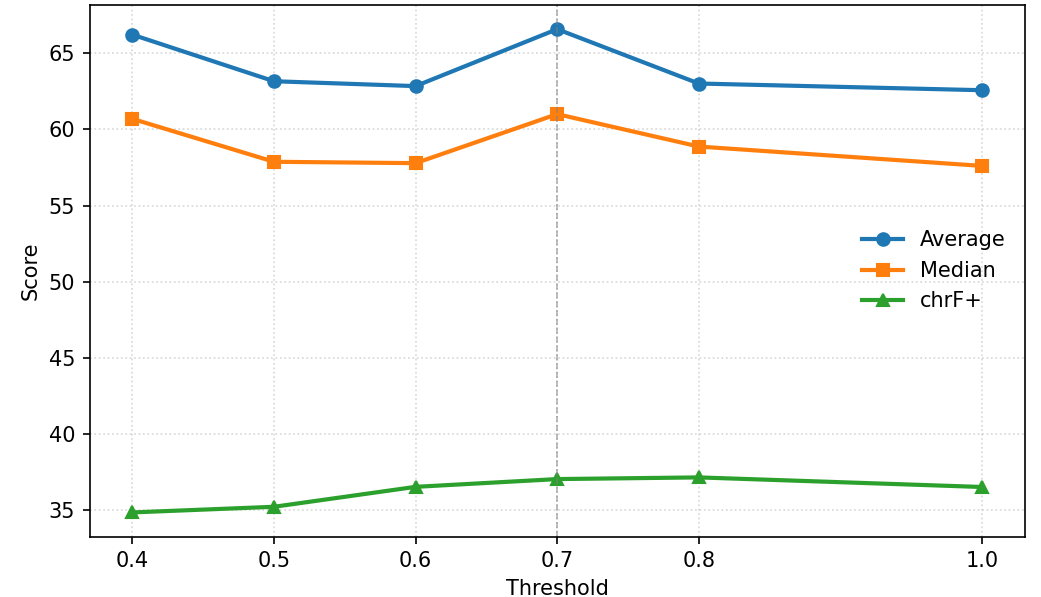}
    \caption{Quality of chosen citation under different similarity thresholds for posthoc citation generation.}
    \label{fig:Similarity}
\end{figure}

\begin{table*}[t]
\section{The quality of verbatim snippets with the gold snippets}
\label{appendix:verbatim}
\centering
\small
\setlength{\tabcolsep}{2pt}
\newcolumntype{Y}{>{\columncolor{asqaYellow}}c}
\newcolumntype{G}{>{\columncolor{eli5Green}}c}
\newcolumntype{P}{>{\columncolor{expertAgua}}c}
\begin{tabular}{l Y|Y|Y G|G|G P|P|P}
\toprule
\multirow{3}{*}{Model}
  & \multicolumn{3}{c}{ASQA}
  & \multicolumn{3}{c}{BioASQ}
  & \multicolumn{3}{c}{ExpertQA} \\
\cmidrule(lr){2-4} \cmidrule(lr){5-7} \cmidrule(lr){8-10}
  & \cellcolor{white}ROUGE-L & \cellcolor{white}Overlap & \cellcolor{white}ChrF++
  & \cellcolor{white}ROUGE-L & \cellcolor{white}Overlap & \cellcolor{white}ChrF++
  & \cellcolor{white}ROUGE-L & \cellcolor{white}Overlap & \cellcolor{white}ChrF++ \\
\midrule
\multicolumn{10}{c}{\textit{Baseline}} \\
\midrule
Qwen3-8B            & 87.93 & 80.63 & 86.38 & 84.88 & 73.44 & 85.92 & 87.70 & 78.60 & 84.61 \\
Gemma-3-12B-it      & 83.05 & 71.36 & 84.76 & 84.45 & 73.11 & 87.90 &  88.35 & 77.34 & 88.54  \\
\midrule
\multicolumn{10}{c}{\textit{Posthoc}} \\
\midrule
FullCite (Qwen3-8B) (ours)           & 92.73 & 88.46 & 93.02 & 81.58 & 68.85 & 89.72 & 91.07 & 86.16 & 89.97 \\
FullCite (Gemma-3-12B-it (ours))     & 87.17 & 77.94 & 89.92 & 81.59 & 71.96 & 89.59 & 88.35 & 77.34 & 88.54 \\
\midrule
\multicolumn{10}{c}{\textit{Constrained}} \\
\midrule
ReClaim             & 64.16 & 45.52 & 47.39 & 61.07 & 41.44 & 44.22 & 56.57 & 37.48 & 43.75 \\
Qwen3-8B (ours)           & 86.30 & 80.32 & 81.17 & 85.39 & 76.81 & 86.22 & 81.35 & 63.59 & 86.71 \\
Gemma-3-12B-it (ours)     & 94.21 & 83.19 & 90.16 & 84.56 & 74.05 & 84.99 & 86.06 & 78.46 & 80.51 \\
\bottomrule
\end{tabular}
\caption{The quality of verbatim snippets with the gold snippets across ASQA, BioASQ, and ExpertQA. \textit{Generate-then-retrieve} strategy attributes the text after generation through retrieval, hence its snippets are always verbatim; therefore, we omit them for this table.}
\label{tab:verbatim}
\end{table*}

\begin{figure*}[ht]
\section{Citation rate per question type}
\label{appendix: citation_rate_qtype}
    \centering
    \includegraphics[width=\linewidth]{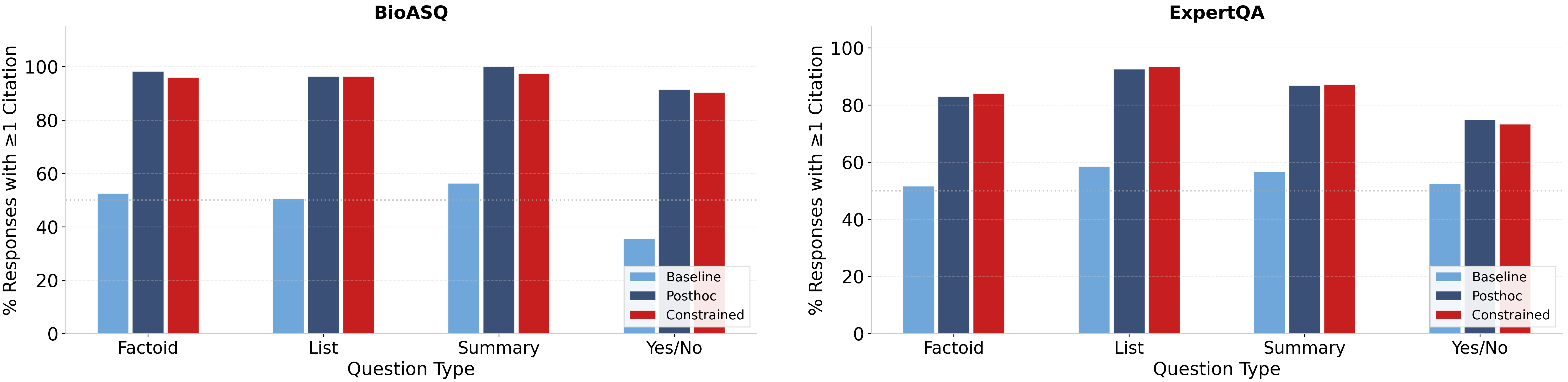}
    \caption{Citation rate for each question type under each setting}
    \label{fig:citation_rate_per_question}
\end{figure*}

\begin{table*}
\section{An example of questions and documents without explicit evidence answers}
\label{appendix:skipped-question-example}
    \centering
    \begin{tabular}{p{3cm}|p{5cm}|p{7cm}}
        Question & Context document & Answer \\
        \hline
        Who won the battle of philippi civil war? & Thomas Armstrong Morris (December 26, 1811 2013 March 23, 1904) was an American railroad executive and civil engineer from Kentucky and a soldier, serving as a brigadier general
of the Indiana Militia in service to the Union during the early months of the American Civil War. & The Battle of Philippi formed part of the Western Virginia Campaign of the American Civil War and was fought in and around Philippi, Virginia (now West Virginia), on June 3, 1861. A Union victory, it was the first organized land action of the war and it brought overnight fame to Maj. Gen. George B. McClellan. Alongside Maj. Gen. McClellan were Thomas A. Morris, a Brigadier General of the Indiana Militia and Isaac Duval, also a Brigadier General. Six Union infantries participated in the battle, including 1st West Virginia Infantry, 2nd West Virginia Infantry (reserve), 6th Indiana Infantry, 7th Indiana Infantry, 9th Indiana Infantry and the 14th Ohio Infantry. \\
\hline
Where is three billboards outside of ebbing missouri filmed? & Three Billboards Outside Ebbing, Missouri is a 2017 crime drama film written, co-produced, and directed by Martin McDonagh and starring Frances McDormand as a Missouri woman who rents three roadside billboards to call attention to her daughter's unsolved rape and murder. &   Principal photography for Three Billboards Outside Ebbing, Missouri began on May 2, 2016 in Sylva, North Carolina and ran for 33 days. Allison Outdoor Advertising of Sylva built the actual billboards, which were put in a pasture near Black Mountain, North Carolina.",
            "The 2017 film Three Billboards Outside of Ebbing, Missouri, was principally filmed in Sylva, North Carolina, while it was filmed in Black Mountain, North Carolina, for the billboard scenes. Sylva is an incorporated town located in central Jackson County in Western North Carolina's Plott Balsam Mountains, USA. Black Mountain is a town in Buncombe County, North Carolina, in the US. \\
    \hline
    \end{tabular}
    \caption{An example from ASQA where the relevant context document misses explicit evidence indicating to the answer to the question. Such examples were excluded.}
    \label{tab:skipped_questions}
\end{table*}
\clearpage

\begin{table*}
\section{An example of questions per each question type}
\label{appendix:data-qtype-example}
    \centering
    \begin{tabular}{p{10cm}|p{2cm}|p{2cm}}
        Question & Question Type & Dataset \\
        \hline
        Please list the short half life Phosphodiesterase 5 inhibitors. & list & BioASQ \\
        Are respiratory measurements correlated with disease severity in Duchenne Muscular Dystrophy? & yes-no & BioASQ \\
        What is the consensus binding site of the TEAD transcription factors? & factoid & BioASQ \\
        What is the role of the receptor tyrosine kinase AXL in malignancy? & summary & BioASQ \\
        \hline
        Which factors strengthen social cohesion within societies? & list & ExpertQA \\
        If my cat ingests oil-based paint, will it get sick? & yes-no & ExpertQA \\
        How many methods exist for children in creative expression? & factoid & ExpertQA \\
        How has the European Court of Justice influenced social mobility within member states through its interpretation and application of European law? & summary & ExpertQA \\
        \hline
        Who has the highest goals in world football? & factoid & ASQA \\
        \hline
    \end{tabular}
    \caption{An example of questions per each question type from each dataset.}
    \label{tab:data-example}
\end{table*}
\clearpage

\begin{table*}
\section{An example of generated evidence snippet}
\label{appendix:snippet-example}
    \centering
    \begin{tabular}{p{7cm}|p{7cm}}
        Context Document & Generated Evidence \\
        \hline
        resources, and dismantle stereotype threats are critically important. Inclusive and culturally responsive learning environments[The teachers] treat us like people with emotions. You have real relationships with your teachers. We want to do our work because we care about our teachers. 
        \textcolor{blue}{Inclusive and culturally \textcolor{red}{responsible} learning environments affirm students value by acknowledging their learning, contributions, and capacity.} 
        It is often said that students learn as much for a teacher as from a teacher. And the teachers they learn the most from are those they believe care about them and see them as worthy of their investment. At Bronxdale High School, staff have developed explicit practices to ensure that they communicate the many ways they value each of their students, including the Affirmation Station shown in Figure 3.2.Figure 3.2 Affirmation Station at Bronxdale High School Source: Ancess, J., Rogers, B., Duncan Grand, D.,  Darling- Hammond, L. (2019). &  \textcolor{blue}{Inclusive and culturally \textcolor{red}{responsive} learning environments affirm students value by acknowledging their learning, contributions, and capacity.} \\
        
    \end{tabular}
    \caption{An example of a generated evidence snippet. Constrained decoding would have marked this output as incorrect at the generation time of \textit{responsive} and would have started from the beginning with an increased temperature. In case of posthoc, it finds the most similar span that fits these token set and assigns the right evidence span because the difference with the original is one word.}
    \label{tab:data-example}
\end{table*}
\clearpage

\begin{table*}[!htbp]
\section{Frequently occurring errors and their statistics}
\label{appendix: error_statistics}
\centering
\footnotesize
\setlength{\tabcolsep}{3pt}
\renewcommand{\arraystretch}{1.1}
\begin{tabular}{llccccccccc}
\toprule
\textbf{Dataset} & \textbf{Setting} 
& \makecell{Avg citation\\ratio \%} 
& \makecell{Full\\coverage \%} 
& \makecell{Doc-only\\cit. \%} 
& \makecell{Refusal\\\%} 
& \makecell{Verbose\\snippet \%} 
& \makecell{Primacy\\pos-1 \%} 
& \makecell{Primacy\\pos-1,2 \%} 
& \makecell{Yes/No\\cit. rate \%} 
& \makecell{Duplication\\rate \%} \\
\midrule
\multirow{3}{*}{BioASQ}    
 & Baseline    &  9.8 &  0.3 & 12.9 &  0.0 &   --  & 55.3 & 81.8 & 35.6 &  0.0 \\
 & Posthoc     & 10.3 &  0.0 &  --  &  3.5 & 14.2 & 54.4 & 82.3 & 91.5 &  8.9 \\
 & Constrained &  9.7 &  0.0 &  --  &  5.0 & 22.2 & 55.2 & 83.9 & 90.4 & 10.3 \\
\midrule
\multirow{3}{*}{ASQA}      
 & Baseline    & 17.0 & 15.3 &  5.3 &  0.2 &   --  & 65.3 & 67.8 &  --  &  0.0 \\
 & Posthoc     & 49.3 & 39.9 &  --  & 41.7 &  5.6 & 64.1 & 65.4 &  --  &  5.3 \\
 & Constrained & 36.1 & 28.2 &  --  & 56.6 & 35.0 & 64.3 & 66.1 &  --  &  5.6 \\
\midrule
\multirow{3}{*}{ExpertQA}  
 & Baseline    & 36.3 & 17.8 &  6.6 &  0.1 &   --  & 36.8 & 53.3 & 52.5 &  0.0 \\
 & Posthoc     & 38.6 & 13.2 &  --  & 17.9 & 13.9 & 50.8 & 64.2 & 74.9 & 25.6 \\
 & Constrained & 38.1 & 13.2 &  --  & 17.6 & 13.4 & 53.2 & 66.2 & 73.4 & 27.1 \\
\bottomrule
\end{tabular}
\caption{Citation behavior metrics across datasets and settings. ``--'' indicates the metric is not applicable for that setting.}
\label{tab:citation-behavior}
\end{table*}

\begin{table*}[t]
\section{Interpretation of rubric for Support and Relevance evaluation}
\label{appendix:rubrics}
\centering
\begin{tabularx}{\linewidth}{c|X}
\hline
Dimension & Scores \\
\hline
\multirow{5}{*}{Support}
  & 1 - No support: The snippet does not support the claim at all; they are unrelated or contradictory \\
  & 2 - Barely support: I see why it was chosen, but I can't say for sure there is support \\
  & 3 - Partial support: The snippet supports some aspects of the claim but not others \\
  & 4 - Support (not all points): Good support, but not all points from the snippet are addressed in the claim \\
  & 5 - Full support: The snippet clearly and completely supports the claim \\
\hline
\multirow{5}{*}{Relevance}
  & 1 - Not relevant: The snippet has nothing to do with the question \\
  & 2 - Barely relevant: Tangentially related but doesn't help answer the question \\
  & 3 - Partially relevant: Addresses some aspect of the question \\
  & 4 - Mostly relevant: Directly related and helpful for answering \\
  & 5 - Fully relevant: Directly answers the question; highly relevant information \\
\hline
\end{tabularx}
\caption{Annotation rubric for the Support and Relevance dimensions.}
\label{tab:support_rubric}
\end{table*}

\begin{figure*}
\section{Citation support and relevance performance over the full set with GPT-5.4}
\label{appendix:llm-eval-scatter}
    \centering
    \includegraphics[width=\linewidth]{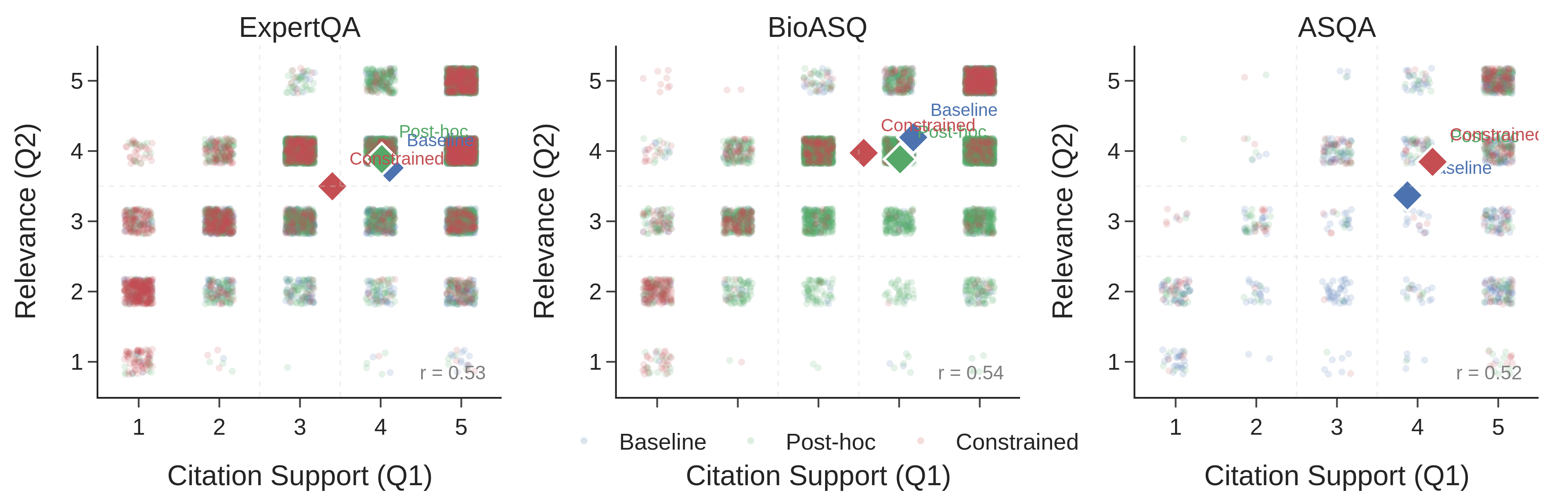}
    \caption{Citation support and relevance performance over the full set with GPT-5.4}
    \label{fig:llm-eval-scatter}
\end{figure*}

        
\FloatBarrier
\end{document}